\documentclass[letterpaper]{article}
\usepackage{spconf,amsmath,graphicx}
\usepackage[caption=false,font=footnotesize]{subfig}

\title{Automatic Image Stylization Using Deep Fully Convolutional Networks}
\name{Feida Zhu \hspace{1.0in} Yizhou Yu}
\address{The University of Hong Kong}
\begin{document}
\ninept
\maketitle
\begin{abstract}
Color and tone stylization strives to enhance unique themes with artistic color and tone adjustments. It has a broad range of applications from professional image postprocessing to photo sharing over social networks. Mainstream photo enhancement softwares provide users with predefined styles, which are often hand-crafted through a trial-and-error process. Such photo adjustment tools lack a semantic understanding of image contents and the resulting global color transform limits the range of artistic styles it can represent. On the other hand, stylistic enhancement needs to apply distinct adjustments to various semantic regions. Such an ability enables a broader range of visual styles. In this paper, we propose a novel deep learning architecture for automatic image stylization, which learns local enhancement styles from image pairs. Our deep learning architecture is an end-to-end deep fully convolutional network performing semantics-aware feature extraction as well as automatic image adjustment prediction. Image stylization can be efficiently accomplished with a single forward pass through our deep network. Experiments on existing datasets for image stylization demonstrate the effectiveness of our deep learning architecture.
\end{abstract}
\section{Introduction}
\label{sec:intro}
Stylistic enhancement adjusts an image for enhancing artistic styles that convey unique themes. Unlike conventional image enhancement focusing on fixing photographic artifacts (under/over exposure, insufficient contrast, etc.), stylistic enhancement involves dramatic color and tone adjustments to achieve distinctive visual effects. For example, the \textit{X-PRO II} filter from mobile photo App Instagram expresses a wistful mood by simulating the cross processing procedure of photographic films. Professional image editing software (such as Adobe Lightroom) and social mobile Apps (such as Instagram) provide users with predefined styles, which are often hand-crafted through a trial-and-error process.

Conventional automatic photo adjustment has difficulty in representing complex color transforms between images before and after adjustment. Most of them merely model global color transforms without considering local semantic contexts. Although more sophisticated adjustments introduce spatially varying effects according to local image statistics, they still lack a semantic understanding of image contents. On the contrary, professional photographers often manually enhance images in a semantics-aware manner.
For instance, when enhancing photos to create a nostalgic theme, photographers might apply more exaggerated adjustments to a photo of \textit{Broadway} taken in year 1950 than a photo of \textit{Burj Khalifa}, which is the tallest skyscraper built in year 2010, as the former is more fitting with the theme.
At a local scale, they employ selection tools to isolate semantic regions (faces, buildings, etc.), which are enhanced with distinct sets of adjustments. For instance, there may exist a demand to apply exaggerated adjustments to foreground objects to help them stand out. The ability of applying distinct adjustments to semantic regions enables a broader range of visual styles.

As stylistic adjustments interact with image semantics and contexts in a complicated manner, it is challenging to manually define the relationships between them. To automatically learn stylistic enhancement from a small set of image exemplars, in this paper, we propose a novel deep learning architecture.
Unlike existing work that integrates hand-crafted features with a small-scale multilayer neural network~\cite{YanZW+16}, our solution is a large-scale deep network. It consists of a fully convolutional network (FCN) for automatic semantic feature extraction and $1$x$1$ convolutional layers for nonlinear transformation and adjustment prediction. Recently, fully convolutional networks~\cite{long2015fully,chen2016semantic,LiYu16} have proven to be efficient and powerful deep learning architectures for image processing and visual understanding tasks, such as semantic image segmentation, contour detection and salient object detection, that need to generate high-resolution outputs. In our deep network, feature maps with sufficiently large receptive fields are computed to model contexts. We further employ 1x1 convolutional layers, which predict color transforms according to contexts and pixel features. We seamlessly integrate the FCN with 1x1 convolutional layers, and an input image can be enhanced with a single forward pass in our deep network.

\begin{figure*}[t]
  \includegraphics[width=1.0\linewidth]{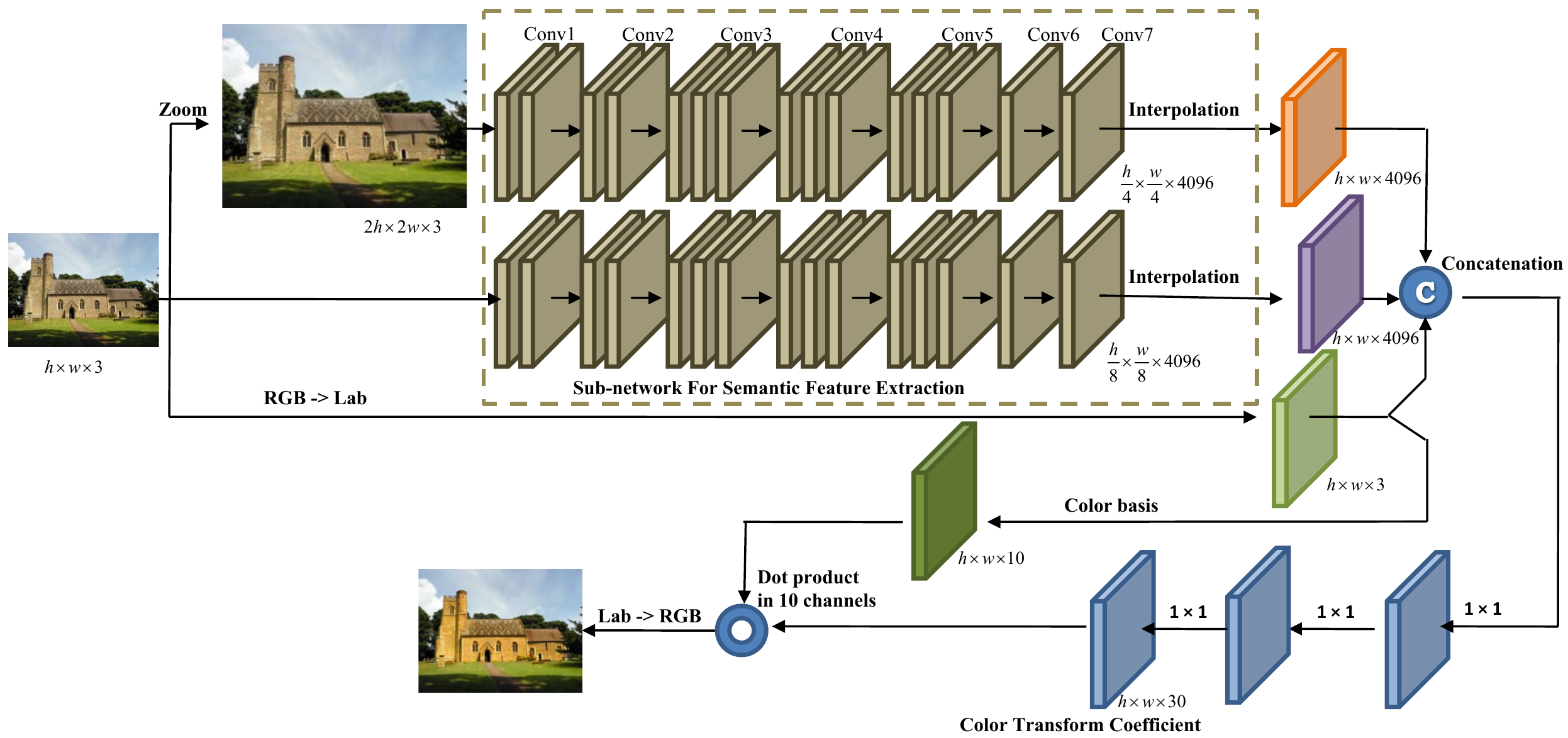}
  \caption{The architecture of our deep neural network for stylistic image enhancement.}
  \label{Fig:architecture}
\end{figure*}

\section{Related Work}
Researchers have made an enormous amount of effort to develop automatic or semi-automatic methods for tone adjustment~\cite{Twoscale_Tone:06}, color management~\cite{wang2010data}, and image smoothing~\cite{jointbilateral,BiHY15}.
In this section, we focus on the most relevant work in data-driven image enhancement.
Data-driven approaches are capable of learning new effects from examples, and thus offer a flexible set of adjustments.
Bychkovsky {\em et al.}~\cite{Bychkovsky} predict global tonal adjustments using a Gaussian process regression model built from a large dataset of images. Their regression model only extracts image global features and does not accommodate semantics-aware local adjustments. Kang {\em et al.}~\cite{PersonalizationSB:2010} introduce user preference in image global enhancement, and retouch a novel image by finding most similar examples in a database and transferring their tone and color adjustments.
Joshi {\em et al.}~\cite{joshi2010personal} retouch imperfect personal photos by leveraging existing high-quality photos of the same person.
Lee {\em et al.}~\cite{LS+16} develop an unsupervised technique for learning content-specific style rankings and transfers highly ranked styles from exemplars to an input photo. However, their styles are still limited to global color and tone transforms.
Wang {\em et al.}~\cite{wang2011example} approximate complex spatially varying tone and color adjustments with piecewise polynomial functions, which rely on low-level image statistics only and are not aware of image semantics.
Yan {\em et al.}~\cite{YanZW+16} present a method for local tone and color adjustments using a combination of image global statistics, contextual semantic features and pixelwise color and spatial features.
Shih {\em et al.} \cite{shih2013data} synthesize images associated with different times of day by learning locally affine models after locating a matching video within a time-lapse video database.
Gatys {\em et al.}~\cite{gatys2016image} perform image style transfer using convolutional neural networks. A new image is synthesized by matching the coarse structures of a content image and the texture features of a style image.


\section{Problem Definition}
Given a set of exemplar image pairs, each representing a photo before and after pixel-level color and tone adjustments following a particular style, we wish to learn a computational model that can automatically adjust a novel input photo in the same style. We cast this learning task as a regression problem as in \cite{YanZW+16}. For completeness, let us first review the original problem definition before presenting our new deep learning based architecture.

We seek a color transformation function $\phi$ such that, for every pixel $p_i$ in the exemplar images, the color transform returned by $\phi$ is $\phi(\theta,x_i)$, which maps the pixel color at $p_i$ before adjustment, $c_i={[L_i ~ a_i ~b_i]}^T$ (CIELab color space), to its corresponding pixel color $y_i$ after adjustment. Here $\theta$ denotes the model parameters and $x_i$ the feature vector at pixel $p_i$. The quadratic color basis $V(c_i)=[L_i^2 ~ a_i^2 ~ b_i^2 ~ L_ia_i ~L_ib_i~ a_ib_i ~L_i~ a_i ~ b_i~1]^T$ is used to absorb high-frequency pixelwise color variations. The product of the color transform $\phi(\theta,x_i)$ and the color basis is a prediction of the enhanced color. Since our color space has 3 channels and the quadratic color basis is a 10-dimensional vector, $\phi(\theta,x_i)$ is in fact a 3x10 matrix. The model parameters of $\theta$ are learnt by minimizing the following objective function, which measures the squared differences between the predicted and groundtruth enhanced colors.
\begin{equation}
\label{eq: cost function}
    \begin{split}
    \arg \min_{\theta} \sum_{i} {||\phi(\theta,x_i)V(c_i)-y_i||}^2
    \end{split}
\end{equation}

\section{Network Architecture}
The architecture of our proposed deep network is shown in Figure~\ref{Fig:architecture}. It relies on automatically learned features at the pixel and contextual levels. The {\em pixel feature} simply consists of all color channels at the considered pixel.
Due to the content-aware nature of our proposed tasks, our network has a fully convolutional sub-network dedicated for extracting contextual semantic features for every pixel in the input image. The final feature map from this sub-network is high-dimensional with 4096 channels, but its resolution is much lower, only one eighth of the resolution of the original input image. The {\em contextual} feature of a pixel consists of two components, which are deep CNN features extracted over two differently sized receptive fields centered at the pixel.

We upsample the input image by a factor of 2, and feed the original and upsampled images into two separate copies of the sub-network for semantic feature computation to extract two feature maps. These feature maps represent semantic contexts at two different scales.
When we compute the feature map from the upsampled image, the size of the receptive field is reduced from $224 \times 224$ to $112 \times 112$. This two-scale scheme is better suited for context modeling. To ensure meaningful features are extracted at image boundaries, we apply reflection padding to the original and upsampled images. Each of these two feature maps then goes through its own {\em interpolation} layer, which makes the feature map reach the full resolution of the original input.

The pixel feature and two-scale contextual feature at every pixel are concatenated and passed through two subsequent 1x1 convolutional layers with nonlinear ReLU activation before being used in the output layer (another 1x1 convolutional layer with 30 channels) to predict per-pixel color transforms. The predicted color transform at pixel $p_i$, $\phi(\theta,x_i)$, is multiplied with its corresponding quadratic color basis vector, $V(c_i)$, in an \textit{enhancement} layer to produce the final adjusted color at $p_i$. Here 1x1 convolutional layers with nonlinear activation are used to perform nonlinear feature transformation. Both pixel features and per-pixel color basis vectors are defined in the CIELab color space. The entire network presented here can be trained or fine-tuned.

Let us now elaborate on the fully convolutional sub-network for semantic feature computation.
Fully convolutional networks are introduced in \cite{long2015fully}, and can be set up by transforming a convolutional neural network (CNN).
The CNN can be pretrained on a large-scale dataset, such as ImageNet~\cite{deng2009imagenet}, to learn discriminative feature representations generically useful for a variety of tasks.
We transform the VGG-16 network~\cite{simonyan2014very} pretrained on ImageNet into a fully convolutional network.
The original VGG-16 network consists of 5 groups of convolutional layers (\textit{conv1}-\textit{conv5}), 5 pooling layers (\textit{pool1}-\textit{pool5}) and 2 fully connected layers (\textit{fc6} and \textit{fc7}). We replace layers \textit{fc6} and \textit{fc7} with 1x1 convolutional layers \textit{conv6} and \textit{conv7} while kernel parameters in \textit{conv6} and \textit{conv7} are copied from \textit{fc6} and \textit{fc7}, respectively. After this change, the spatial dimensions of the feature map at the deepest layer \textit{conv7} is reduced by a factor of 32 due to the 5 pooling layers. Then the stride of pooling kernels in layers \textit{pool4} and \textit{pool5} is reduced from $2$ to $1$. Thus the spatial resolution of feature maps is not reduced across these two layers. As a result, the overall spatial resolution is merely reduced by a factor of 8. Feature maps with high spatial resolutions have been shown to be vital in attaining superior performance in image processing~\cite{long2015fully}.
Changing the kernel stride at a pooling layer alone makes the following convolutional layer have a different receptive field and extract meaningless features.
We use \textit{dilated convolution}~\cite{yu2015multi} to increase the input stride of feature maps for layers \textit{conv5} and \textit{conv6} by factors of 2 and 4 respectively to compensate the change of pooling kernels in \textit{pool4} and \textit{pool5}. To obtain pixelwise semantic features, we fine-tune the above transformed network on the ADE20K dataset~\cite{ADE20Kdataset} with 150 object categories. The ADE20K dataset is specifically designed for scene parsing.
Finally, if an input image of size $H \times W \times 3$ is sent into this fine-tuned FCN, a feature map of size $\frac{H}{8} \times \frac{W}{8} \times 4096$ is extracted from layer \textit{conv7}.


There exist significant qualitative differences between our new deep learning based framework and the method in~\cite{YanZW+16}. First, our proposed framework here extracts semantic features automatically learned from raw data using a fully convolutional neural network while the method in \cite{YanZW+16} only makes use of traditional neural networks and hand-crafted features (e.g. intensity distribution, label histograms, etc). Automatically learned features are much more powerful than hand-crafted features.
Second, our entire deep network is end-to-end trainable and final results are predicted at the pixel level while the method in \cite{YanZW+16} can only train fully connected layers and final results are predicted at the superpixel level. Thus the method in \cite{YanZW+16} does not fully exploit the power of deep learning.
Third, to form a contextual feature descriptor for reflecting the local semantics and appearance around a pixel or superpixel, Yan  {\em et al.} \cite{YanZW+16} rely on a traditional scene parser~\cite{scene_parsing} to label background image regions, and a set of cascaded object detectors \cite{wang2013regionlets} to detect and classify foreground objects into a small set of predefined categories. By merging scene parsing results and object detections into a semantic label map, they compute label histograms using a multiscale spatial pooling scheme. However, both the scene parser and the object detectors they use have been substantially outperformed by recent deep learning models~\cite{long2015fully,ren2015faster,liu2015ssd,chen2016semantic}. Furthermore, in contrast to the continuous deep CNN features used in our contextual features, their category-oriented discrete label map is more prone to quantization errors, which can lead to severe artifacts in the final results.
Fourth, the fully convolutional network in our proposed new framework has an efficient GPU implementation. CNN feature computation in our new framework can be completed on the order of one second. In contrast, the scene parser and object detectors used in \cite{YanZW+16} are two orders of magnitude slower.

\begin{figure*}[!t]
  \centerline{
    \includegraphics[width=0.18\textwidth]{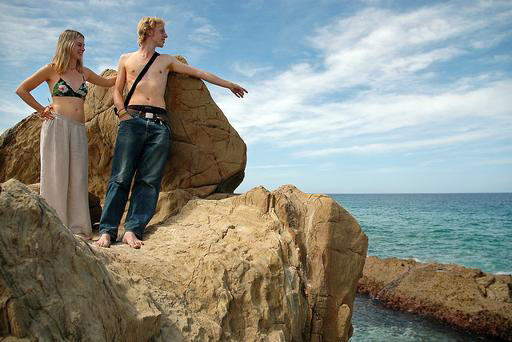}\hfill
    \includegraphics[width=0.18\textwidth]{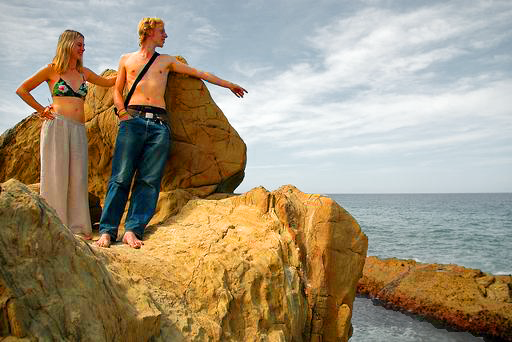}\hfill
    \includegraphics[width=0.18\textwidth]{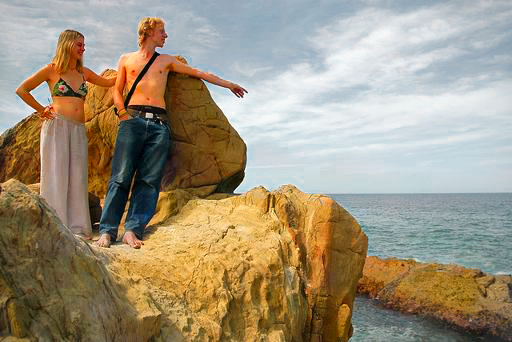}\hfill
    \includegraphics[width=0.18\textwidth]{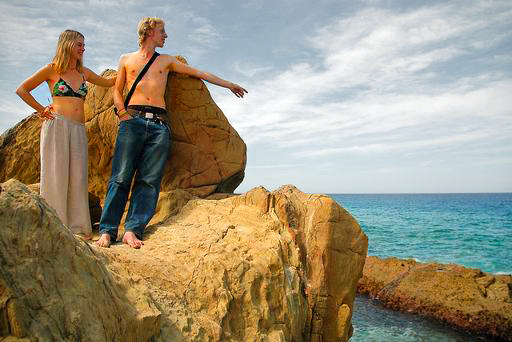}\hfill
    \includegraphics[width=0.216\textwidth]{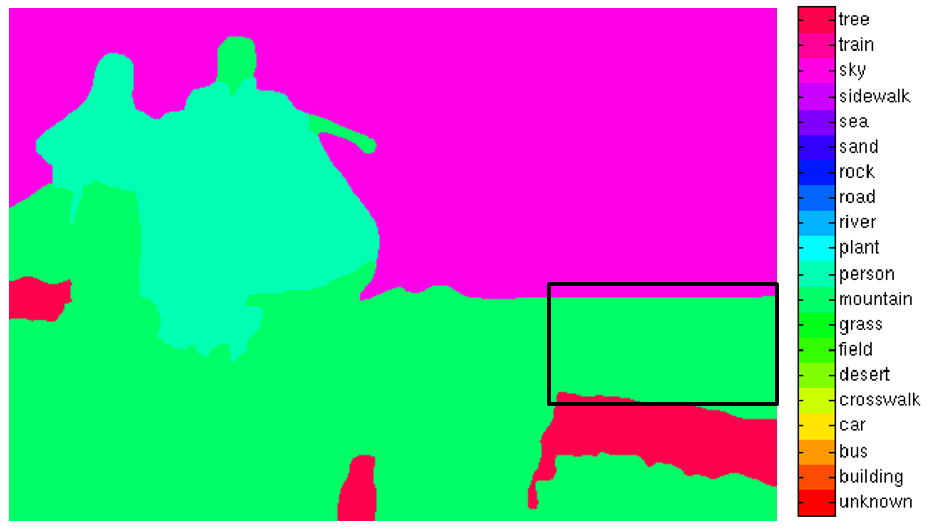}
  }
  \centerline{
    \includegraphics[width=0.18\textwidth]{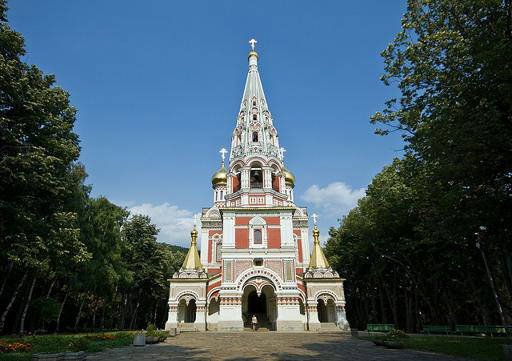}\hfill
    \includegraphics[width=0.18\textwidth]{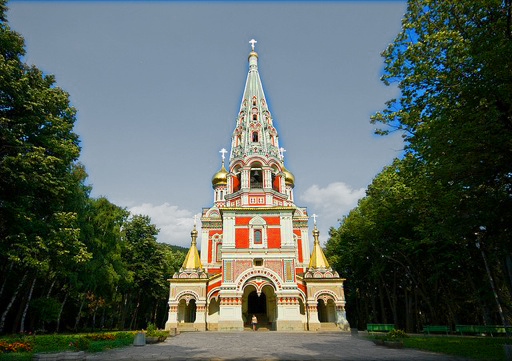}\hfill
    \includegraphics[width=0.18\textwidth]{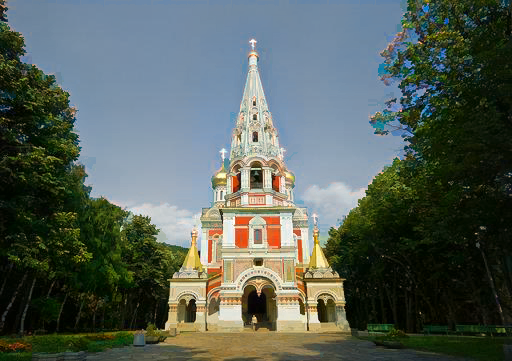}\hfill
    \includegraphics[width=0.18\textwidth]{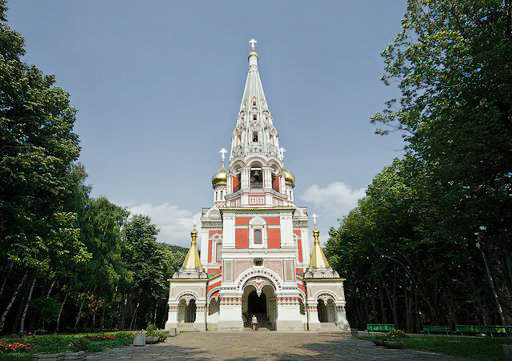}\hfill
    \includegraphics[width=0.216\textwidth]{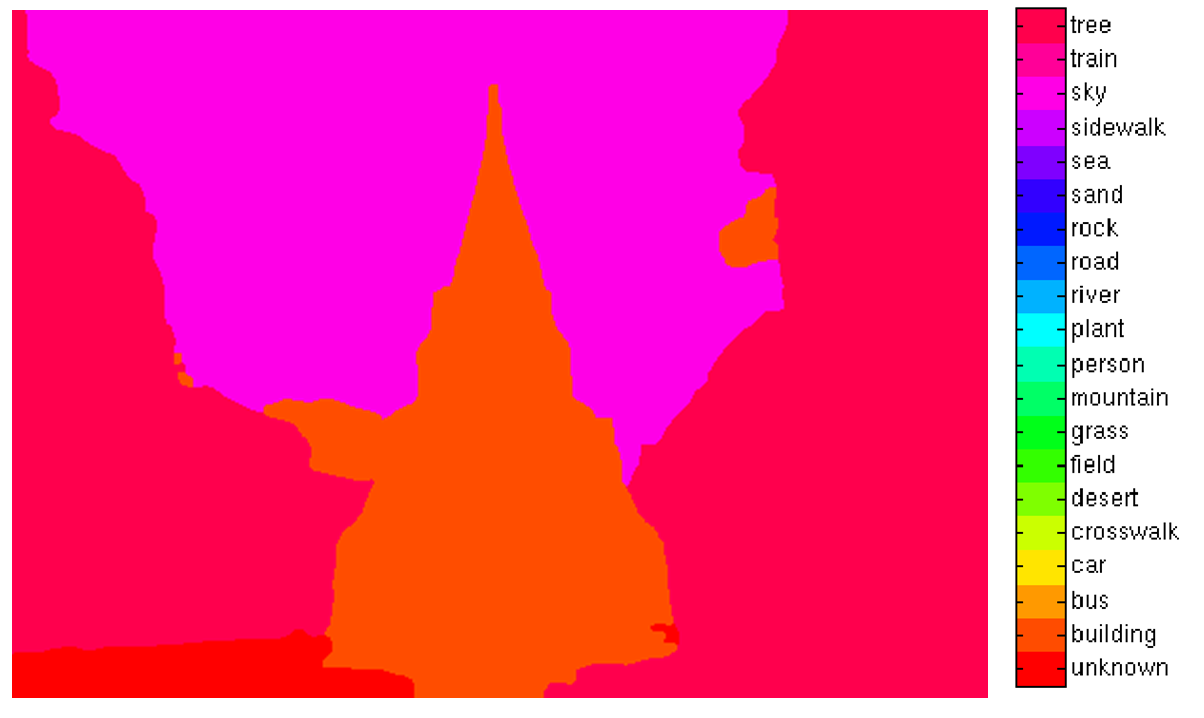}
  }
  \centerline{\small \hfill Input Image \hfill\hfill Ground Truth \hfill\hfill Our Result \hfill\hfill Result from \protect\cite{YanZW+16} \hfill Scene Labels from \protect\cite{YanZW+16}
  }
  \caption{Comparisons with the method in \protect{\cite{YanZW+16}} on examples from the Foreground Popout style. In the top row, their method mislabels `sea' as `mountain' and the saturation of this mislabeled region is incorrectly increased. In the bottom row, although their method labels the `building' region correctly, it still adjusts its color incorrectly, which reveals the limitation of their feature description. In contrast, our proposed method produces much more robust visual results by using contextual semantic features extracted using deep fully convolutional networks.}
\label{fig:comp-yan}
\end{figure*}

\section{Experimental Results}
\subsection{Results on the Uniform Dataset}
Here we report image stylization results from our deep network on the Uniform dataset introduced in \cite{YanZW+16}. This dataset includes 115 images and three local enhancement styles. In the following, we briefly review these three styles.

The first style, ``\textbf{Foreground Pop-out}", increases the contrast and color saturation of foreground objects while decreasing the color saturation of background. In general, this style makes the foreground objects more salient and colorful while deemphasizing the background. The second style, ``\textbf{Local Xpro}", is more complex compared to the previous effect. It generalizes the popular ``cross processing" effect in a local manner. When creating this style, a photographer first defined multiple profiles in Photoshop, each of which is specifically tailored for a subset of semantic categories. All the profiles share a common series of operations, such as hue/saturation adjustment and brightness/contrast manipulation. Nonetheless, each profile defines a distinct set of adjustment parameters tailored for its corresponding semantic categories. Although the profiles roughly follow the ``cross processing" style, the choice of local profiles and additional minor image editing operations were heavily influenced by the photographer's personal taste which can be naturally learned through exemplars. The third style, ``\textbf{Watercolor}", gives viewers artistic impressions. It creates ``brush" effects over the input image. All pixels inside the region covered by a single brush stroke share the same color. This mimics the ``watercolor" painting style. This style also gives rise to complex spatially varying color adjustments.

We have trained our entire deep network from end to end to learn all three styles from this dataset. As in \cite{YanZW+16}, for each style, 70 images are used for training and the remaining 45 images are used for testing.
We have calculated the mean per-pixel $L^2$ distance in the CIELab color space between our enhanced images and the groundtruth results as well as between the input images and the groundtruth results. They are shown in Table~\ref{table:local compare}.

\begin{table}[!bt]
\centering
\caption{Mean per-pixel $L^2$ distances between input images and groundtruth adjusted images and mean per-pixel $L^2$ testing errors between automatically adjusted results and groundtruth results.}
\label{table:local compare}
\resizebox{0.5\textwidth}{!}
{
\renewcommand{\arraystretch}{1.5}
\begin{tabular}{|l|l|l|l|l|}
\hline
Style              & \begin{tabular}[l]{@{}l@{}}Groundtruth\\ $L^2$ distance\end{tabular} & Yan et al. \cite{YanZW+16} & Our Method \\ \hline
Foreground Pop-out & 13.86                                                                 & 7.08                            & 7.10       \\ \hline
Local Xpro         & 19.71                                                                 & 7.43                            & 6.98       \\ \hline
Watercolor         & 15.30                                                                 & 7.20                            & 6.88       \\ \hline
\end{tabular}
}
\end{table}

We primarily compare our proposed method here against the one in \cite{YanZW+16} because the method in \cite{YanZW+16} is the most recent work capable of performing spatially varying photo adjustment. It has been demonstrated that their method outperforms all earlier relevant techniques. Comparison of numerical errors on testing images are shown in the third and fourth columns of Table~\ref{table:local compare}.
Our method achieves much lower numerical errors on the Local Xpro and Watercolor styles, and comparable errors on the Foreground Pop-out style.

Two visual comparisons are shown in Fig~\ref{fig:comp-yan} on the {\em Foreground Pop-out} style.
First, contextual features in \cite{YanZW+16} are built upon hand-crafted features and discrete label maps, whose accuracy and reliability cannot match the performance of continuous deep features computed in the proposed method. Therefore, their stylized results are more likely to have artifacts due to incorrect labeling. One such example from the Foreground Popout style is shown in the top row of Fig~\ref{fig:comp-yan} , where `sea' is mislabeled as `mountain' with their method and the saturation of this mislabeled region is incorrectly increased.
Second, their method requires the definition of a set of semantic categories. When this set is limited (20 categories used in \cite{YanZW+16}), their contextual feature would not work well on testing images containing object categories beyond this predefined set. Another example from the Foreground Popout style is shown in the bottom row of Fig~\ref{fig:comp-yan}, where the saturation of `building' is not increased with their method even though it is labeled correctly. In contrast, contextual features extracted using our new FCN based architecture tend to be more robust.

\begin{table}[bt]
\centering
\caption{Statistics of Cold and Spirng styles. Middle: mean per-pixel $L^2$ distance between original images and the ground truth. Right: mean per-pixel $L^2$ distance between our enhanced results and the ground truth.}
\label{table: spring cold}
{
\renewcommand{\arraystretch}{1.5}
\begin{tabular}{|l|l|l|}
\hline
Style & \begin{tabular}[c]{@{}l@{}}Ground Truth ($L^2$ distance) \end{tabular} & Our Method \\ \hline
Cold   & 16.95                                                                 & 1.98       \\ \hline
Spring & 18.56                                                                 & 3.56       \\ \hline
\end{tabular}
}
\end{table}

\begin{figure}[!bt]
\captionsetup[subfigure]
  { labelformat=empty}
\centering
\subfloat[Input image]{\includegraphics[width=0.15\textwidth]{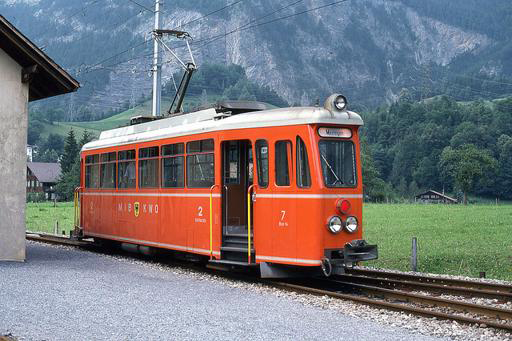}}~
\subfloat[Spring effect]{\includegraphics[width=0.15\textwidth]{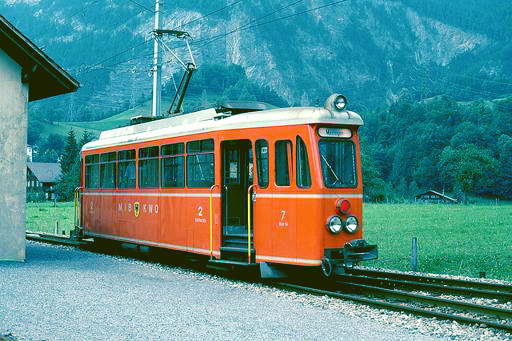}}~
\subfloat[Our result]{\includegraphics[width=0.15\textwidth]{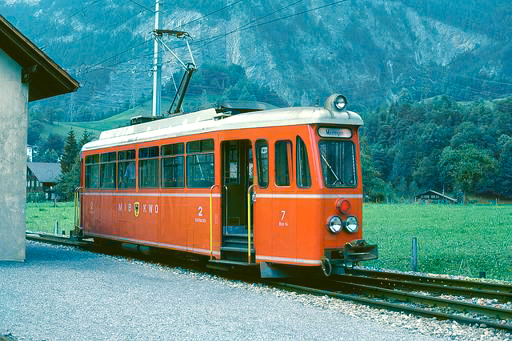}}\\
\subfloat[Input image]{\includegraphics[width=0.15\textwidth]{figs/4677364341_58a614bf9d_z-orig.png}}~
\subfloat[Cold effect]{\includegraphics[width=0.15\textwidth]{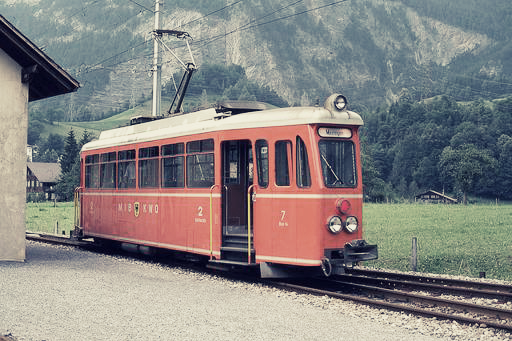}}~
\subfloat[Our result]{\includegraphics[width=0.15\textwidth]{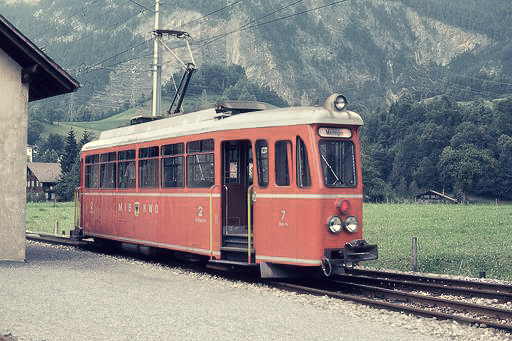}}
\caption{Examples of global `Spring' and `Cold' styles. \textbf{Left}: Input image. \textbf{Middle}: Ground truth. \textbf{Right}: Our result.}
\label{fig: spring cold}
\end{figure}

\subsection{Global Styles}
Our deep network can readily learn global image adjustment styles as well. We asked a photographer to create additional stylistic effects which are saved as ``action" files in Photoshop, each of which contains a series of operations such as saturation adjustment, tone adjustment, and curve tuning. An ``action" is globally applied to the original images in the Uniform dataset~\cite{YanZW+16} to form a set of image exemplars for a global enhancement style. It is also convenient for us to obtain from the Internet various stylistic ``action" files shared by photo retouching enthusiasts. Here we demonstrate model training and testing for two global effects as examples.

The first global effect (called ``Spring") gives viewers the feeling of vitality, and the second global effect (called ``Cold") tends to make a photo look vintage. Fig~\ref{fig: spring cold} shows enhancement examples in these two styles.
The mean per-pixel $L^2$ distance between the original images and their groundtruth enhanced images reaches 16.95 and 18.56 for the ``Cold" and ``Spring" styles, respectively. This indicates that these two effects make significant color changes to the original images. The small testing errors of our trained models shown in Table~\ref{table: spring cold} numerically demonstrate the strength of our method in learning such global effects. Comparing the middle column and the right column in Fig~\ref{fig: spring cold}, we can see that the enhanced results from our model look nearly the same as the ground truth.

\section{Conclusions}
In this paper, we have presented a novel deep learning architecture for exemplar-based image stylization, which learns local enhancement styles from image pairs. Our deep learning architecture is an end-to-end deep fully convolutional network performing semantics-aware feature extraction as well as automatic image adjustment prediction. Image stylization can be efficiently accomplished with a single forward pass through our deep network. Experiments on existing datasets for image stylization demonstrate the effectiveness of our deep network.

\bibliographystyle{IEEEbib}
\bibliography{ref}

\end{document}


\onecolumn
\centerline{\LARGE \bf AUTOMATIC IMAGE STYLIZATION USING DEEP FULLY CONVOLUTIONAL NETWORKS}
\vspace{4mm}

\centerline{(Supplemental Materials)}
\vspace{20mm}

\noindent \textbf{\Large Content Outline}
\vspace{2mm}

\hyperref[sec:popout]{$\bullet$ \textbf{\large Foreground Popout effect}}
\vspace{2mm}

\hyperref[sec:xpro]{$\bullet$ \textbf{\large Local Xpro effect}}
\vspace{2mm}

\hyperref[sec:watercolor]{$\bullet$ \textbf{\large Watercolor effect}}
\vspace{2mm}

\hyperref[sec:compare1]{$\bullet$ \textbf{\large Comparison with [Yan {\em et al.} 2016] on Foreground Popout effect}}
\vspace{2mm}

\hyperref[sec:compare2]{$\bullet$ \textbf{\large Comparison with [Yan {\em et al.} 2016] on Local Xpro effect}}
\vspace{2mm}

\hyperref[sec:compare3]{$\bullet$ \textbf{\large Comparison with [Yan {\em et al.} 2016] on Watercolor effect}}
\vspace{2mm}

\hyperref[sec:cold]{$\bullet$ \textbf{\large Cold effect}}
\vspace{2mm}

\hyperref[sec:spring]{$\bullet$ \textbf{\large Spring effect}}

\clearpage

\section{Foreground Popout effect}
\label{sec:popout}


\begin{figure}[H]
\captionsetup[subfigure]{labelformat=empty}
\centering
\subfloat{\includegraphics[width=0.3\textwidth]{orig_images/0087.png}}~
\subfloat{\includegraphics[width=0.3\textwidth]{groundtruth/popout_images/0087.png}}~
\subfloat{\includegraphics[width=0.3\textwidth]{enhanced/popout_images/0087.png}}\\
\subfloat{\includegraphics[width=0.3\textwidth]{orig_images/0092.png}}~
\subfloat{\includegraphics[width=0.3\textwidth]{groundtruth/popout_images/0092.png}}~
\subfloat{\includegraphics[width=0.3\textwidth]{enhanced/popout_images/0092.png}}\\
\subfloat{\includegraphics[width=0.3\textwidth]{orig_images/0095.png}}~
\subfloat{\includegraphics[width=0.3\textwidth]{groundtruth/popout_images/0095.png}}~
\subfloat{\includegraphics[width=0.3\textwidth]{enhanced/popout_images/0095.png}}\\
\subfloat{\includegraphics[width=0.3\textwidth]{orig_images/0071.png}}~
\subfloat{\includegraphics[width=0.3\textwidth]{groundtruth/popout_images/0071.png}}~
\subfloat{\includegraphics[width=0.3\textwidth]{enhanced/popout_images/0071.png}}\\
\subfloat{\includegraphics[width=0.3\textwidth]{orig_images/0113.png}}~
\subfloat{\includegraphics[width=0.3\textwidth]{groundtruth/popout_images/0113.png}}~
\subfloat{\includegraphics[width=0.3\textwidth]{enhanced/popout_images/0113.png}}\\
\caption{Examples of Foreground Popout effect.
\textbf{First column:} Input image. \textbf{Second column:} Ground truth.
\textbf{Third column:} Our result.}
\label{fig:popout2}
\end{figure}

\section{Local Xpro effect}
\label{sec:xpro}


\begin{figure}[H]
\captionsetup[subfigure]{labelformat=empty}
\centering
\subfloat{\includegraphics[width=0.3\textwidth]{orig_images/0089.png}}~
\subfloat{\includegraphics[width=0.3\textwidth]{groundtruth/xpro_images/0089.png}}~
\subfloat{\includegraphics[width=0.3\textwidth]{enhanced/xpro_images/0089.png}}\\
\subfloat{\includegraphics[width=0.3\textwidth]{orig_images/0093.png}}~
\subfloat{\includegraphics[width=0.3\textwidth]{groundtruth/xpro_images/0093.png}}~
\subfloat{\includegraphics[width=0.3\textwidth]{enhanced/xpro_images/0093.png}}\\
\subfloat{\includegraphics[width=0.3\textwidth]{orig_images/0097.png}}~
\subfloat{\includegraphics[width=0.3\textwidth]{groundtruth/xpro_images/0097.png}}~
\subfloat{\includegraphics[width=0.3\textwidth]{enhanced/xpro_images/0097.png}}\\
\subfloat{\includegraphics[width=0.3\textwidth]{orig_images/0100.png}}~
\subfloat{\includegraphics[width=0.3\textwidth]{groundtruth/xpro_images/0100.png}}~
\subfloat{\includegraphics[width=0.3\textwidth]{enhanced/xpro_images/0100.png}}\\
\subfloat{\includegraphics[width=0.3\textwidth]{orig_images/0104.png}}~
\subfloat{\includegraphics[width=0.3\textwidth]{groundtruth/xpro_images/0104.png}}~
\subfloat{\includegraphics[width=0.3\textwidth]{enhanced/xpro_images/0104.png}}\\
\caption{Examples of Local Xpro effect.
\textbf{First column:} Input image. \textbf{Second column:} Ground truth.
\textbf{Third column:} Our result.}
\label{fig:xpro2}
\end{figure}

\section{Watercolor effect}
\label{sec:watercolor}


\begin{figure}[H]
\captionsetup[subfigure]{labelformat=empty}
\centering
\subfloat{\includegraphics[width=0.3\textwidth]{orig_images/0074.png}}~
\subfloat{\includegraphics[width=0.3\textwidth]{groundtruth/watercolor_images/0074.png}}~
\subfloat{\includegraphics[width=0.3\textwidth]{enhanced/watercolor_images/0074.png}}\\
\subfloat{\includegraphics[width=0.3\textwidth]{orig_images/0090.png}}~
\subfloat{\includegraphics[width=0.3\textwidth]{groundtruth/watercolor_images/0090.png}}~
\subfloat{\includegraphics[width=0.3\textwidth]{enhanced/watercolor_images/0090.png}}\\
\subfloat{\includegraphics[width=0.3\textwidth]{orig_images/0082.png}}~
\subfloat{\includegraphics[width=0.3\textwidth]{groundtruth/watercolor_images/0082.png}}~
\subfloat{\includegraphics[width=0.3\textwidth]{enhanced/watercolor_images/0082.png}}\\
\subfloat{\includegraphics[width=0.3\textwidth]{orig_images/0101.png}}~
\subfloat{\includegraphics[width=0.3\textwidth]{groundtruth/watercolor_images/0101.png}}~
\subfloat{\includegraphics[width=0.3\textwidth]{enhanced/watercolor_images/0101.png}}\\
\subfloat{\includegraphics[width=0.3\textwidth]{orig_images/0115.png}}~
\subfloat{\includegraphics[width=0.3\textwidth]{groundtruth/watercolor_images/0115.png}}~
\subfloat{\includegraphics[width=0.3\textwidth]{enhanced/watercolor_images/0115.png}}\\
\caption{Examples of Watercolor effect.
\textbf{First column:} Input image. \textbf{Second column:} Ground truth.
\textbf{Third column:} Our result.}
\label{fig:watercolor2}
\end{figure}

\section{Comparison with [Yan {\em et al.} 2016] on Foreground Popout effect}
\label{sec:compare1}

\begin{figure}[H]
\captionsetup[subfigure]{labelformat=empty}
\centering
\subfloat{\includegraphics[width=0.24\textwidth]{orig_images/0095.png}}~
\subfloat{\includegraphics[width=0.24\textwidth]{groundtruth/popout_images/0095.png}}~
\subfloat{\includegraphics[width=0.24\textwidth]{enhanced/popout_images/0095.png}}~
\subfloat{\includegraphics[width=0.24\textwidth]{enhanced_by_zhicheng/popout_images/0095.png}}\\
\subfloat{\includegraphics[width=0.24\textwidth]{orig_images/0071.png}}~
\subfloat{\includegraphics[width=0.24\textwidth]{groundtruth/popout_images/0071.png}}~
\subfloat{\includegraphics[width=0.24\textwidth]{enhanced/popout_images/0071.png}}~
\subfloat{\includegraphics[width=0.24\textwidth]{enhanced_by_zhicheng/popout_images/0071.png}}\\
\subfloat{\includegraphics[width=0.24\textwidth]{orig_images/0113.png}}~
\subfloat{\includegraphics[width=0.24\textwidth]{groundtruth/popout_images/0113.png}}~
\subfloat{\includegraphics[width=0.24\textwidth]{enhanced/popout_images/0113.png}}~
\subfloat{\includegraphics[width=0.24\textwidth]{enhanced_by_zhicheng/popout_images/0113.png}}\\
\caption{Comparison with [Yan {\em et al.} 2016] on Foreground Popout effect.
\textbf{First column:} Input image. \textbf{Second column:} Ground truth.
\textbf{Third column:} Our result. \textbf{Fourth column:} result of [Yan {\em et al.} 2016] }
\label{fig:watercolor1}
\end{figure}

\clearpage
\section{Comparison with [Yan {\em et al.} 2016] on Local Xpro effect}
\label{sec:compare2}

\begin{figure}[H]
\captionsetup[subfigure]{labelformat=empty}
\centering
\subfloat{\includegraphics[width=0.24\textwidth]{orig_images/0088.png}}~
\subfloat{\includegraphics[width=0.24\textwidth]{groundtruth/xpro_images/0088.png}}~
\subfloat{\includegraphics[width=0.24\textwidth]{enhanced/xpro_images/0088.png}}~
\subfloat{\includegraphics[width=0.24\textwidth]{enhanced_by_zhicheng/xpro_images/0088.png}}\\
\subfloat{\includegraphics[width=0.24\textwidth]{orig_images/0104.png}}~
\subfloat{\includegraphics[width=0.24\textwidth]{groundtruth/xpro_images/0104.png}}~
\subfloat{\includegraphics[width=0.24\textwidth]{enhanced/xpro_images/0104.png}}~
\subfloat{\includegraphics[width=0.24\textwidth]{enhanced_by_zhicheng/xpro_images/0104.png}}\\
\caption{Comparison with [Yan {\em et al.} 2016] on Local Xpro effect.
\textbf{First column:} Input image. \textbf{Second column:} Ground truth.
\textbf{Third column:} Our result. \textbf{Fourth column:} result of [Yan {\em et al.} 2016] }
\label{fig:watercolor1}
\end{figure}

\section{Comparison with [Yan {\em et al.} 2016] on Watercolor effect}
\label{sec:compare3}

\begin{figure}[H]
\captionsetup[subfigure]{labelformat=empty}
\centering
\subfloat{\includegraphics[width=0.24\textwidth]{orig_images/0101.png}}~
\subfloat{\includegraphics[width=0.24\textwidth]{groundtruth/watercolor_images/0101.png}}~
\subfloat{\includegraphics[width=0.24\textwidth]{enhanced/watercolor_images/0101.png}}~
\subfloat{\includegraphics[width=0.24\textwidth]{enhanced_by_zhicheng/watercolor_images/0101.png}}\\
\subfloat{\includegraphics[width=0.24\textwidth]{orig_images/0077.png}}~
\subfloat{\includegraphics[width=0.24\textwidth]{groundtruth/watercolor_images/0077.png}}~
\subfloat{\includegraphics[width=0.24\textwidth]{enhanced/watercolor_images/0077.png}}~
\subfloat{\includegraphics[width=0.24\textwidth]{enhanced_by_zhicheng/watercolor_images/0077.png}}\\
\caption{Comparison with [Yan {\em et al.} 2016] on Watercolor effect.
\textbf{First column:} Input image. \textbf{Second column:} Ground truth.
\textbf{Third column:} Our result. \textbf{Fourth column:} result of [Yan {\em et al.} 2016] }
\label{fig:watercolor1}
\end{figure}

\clearpage
\section{Cold effect}
\label{sec:cold}

\begin{figure}[H]
\captionsetup[subfigure]{labelformat=empty}
\subfloat{\includegraphics[width=0.3\textwidth]{orig_images/0071.png}}~
\subfloat{\includegraphics[width=0.3\textwidth]{groundtruth/cold_images/0071.png}}~
\subfloat{\includegraphics[width=0.3\textwidth]{enhanced/cold_images/0071.png}}\\
\subfloat{\includegraphics[width=0.3\textwidth]{orig_images/0114.png}}~
\subfloat{\includegraphics[width=0.3\textwidth]{groundtruth/cold_images/0114.png}}~
\subfloat{\includegraphics[width=0.3\textwidth]{enhanced/cold_images/0114.png}}\\
\subfloat{\includegraphics[width=0.3\textwidth]{orig_images/0113.png}}~
\subfloat{\includegraphics[width=0.3\textwidth]{groundtruth/cold_images/0113.png}}~
\subfloat{\includegraphics[width=0.3\textwidth]{enhanced/cold_images/0113.png}}\\
\subfloat{\includegraphics[width=0.3\textwidth]{orig_images/0112.png}}~
\subfloat{\includegraphics[width=0.3\textwidth]{groundtruth/cold_images/0112.png}}~
\subfloat{\includegraphics[width=0.3\textwidth]{enhanced/cold_images/0112.png}}\\
\subfloat{\includegraphics[width=0.3\textwidth]{orig_images/0111.png}}~
\subfloat{\includegraphics[width=0.3\textwidth]{groundtruth/cold_images/0111.png}}~
\subfloat{\includegraphics[width=0.3\textwidth]{enhanced/cold_images/0111.png}}\\
\caption{Examples of Cold effect.
\textbf{First column:} Input image. \textbf{Second column:} Ground truth.
\textbf{Third column:} Our result.}
\label{fig:popout1}
\end{figure}

\clearpage

\section{Spring effect}
\label{sec:spring}

\begin{figure}[H]
\captionsetup[subfigure]{labelformat=empty}
\centering
\subfloat{\includegraphics[width=0.3\textwidth]{orig_images/0095.png}}~
\subfloat{\includegraphics[width=0.3\textwidth]{groundtruth/spring_images/0095.png}}~
\subfloat{\includegraphics[width=0.3\textwidth]{enhanced/spring_images/0095.png}}\\
\subfloat{\includegraphics[width=0.3\textwidth]{orig_images/0096.png}}~
\subfloat{\includegraphics[width=0.3\textwidth]{groundtruth/spring_images/0096.png}}~
\subfloat{\includegraphics[width=0.3\textwidth]{enhanced/spring_images/0096.png}}\\
\subfloat{\includegraphics[width=0.3\textwidth]{orig_images/0097.png}}~
\subfloat{\includegraphics[width=0.3\textwidth]{groundtruth/spring_images/0097.png}}~
\subfloat{\includegraphics[width=0.3\textwidth]{enhanced/spring_images/0097.png}}\\
\subfloat{\includegraphics[width=0.3\textwidth]{orig_images/0098.png}}~
\subfloat{\includegraphics[width=0.3\textwidth]{groundtruth/spring_images/0098.png}}~
\subfloat{\includegraphics[width=0.3\textwidth]{enhanced/spring_images/0098.png}}\\
\subfloat{\includegraphics[width=0.3\textwidth]{orig_images/0099.png}}~
\subfloat{\includegraphics[width=0.3\textwidth]{groundtruth/spring_images/0099.png}}~
\subfloat{\includegraphics[width=0.3\textwidth]{enhanced/spring_images/0099.png}}\\
\caption{Examples of Spring effect.
\textbf{First column:} Input image. \textbf{Second column:} Ground truth.
\textbf{Third column:} Our result.}
\label{fig:popout1}
\end{figure}

\clearpage